\documentclass[conference]{IEEEtran}
\IEEEoverridecommandlockouts
\usepackage{cite}
\usepackage{amsmath,amssymb,amsfonts}
\usepackage{algorithmic}
\usepackage{graphicx}
\usepackage{textcomp}
\usepackage{xcolor}
\usepackage{caption}
\usepackage{subcaption}
\usepackage{booktabs}
\usepackage{comment}
\usepackage{float}
\usepackage{hyperref}

\begin{document}

\title{Building Footprint Extraction in Dense Areas using
Super Resolution and Frame Field Learning}

\author{\IEEEauthorblockN{1\textsuperscript{st} Vuong Nguyen}
\IEEEauthorblockA{\textit{Department of Computer Science} \\
\textit{University of Houston}\\
Houston, Texas, USA \\
dnguy222@cougarnet.uh.edu}
\and
\IEEEauthorblockN{2\textsuperscript{nd} Trong-Anh Ho}
\IEEEauthorblockA{\textit{Skymap Global} \\
Hanoi, Vietnam \\
hotronganh95@gmail.com}
\and
\IEEEauthorblockN{3\textsuperscript{rd} Duc-Anh Vu}
\IEEEauthorblockA{\textit{Nanyang Technological University} \\
Singapore \\
ducanh001@e.ntu.edu.sg}
\and
\IEEEauthorblockN{4\textsuperscript{th} Nguyen Thi Ngoc Anh}
\IEEEauthorblockA{\textit{School of Applied Mathematics and Informatics} \\
\textit{Hanoi University of Science and Technology}\\
Hanoi, Vietnam \\
anh.nguyenthingoc@hust.edu.vn}
\and
\IEEEauthorblockN{5\textsuperscript{th} Tran Ngoc Thang$^*$}
\IEEEauthorblockA{\textit{School of Applied Mathematics and Informatics} \\
\textit{Hanoi University of Science and Technology}\\
Hanoi, Vietnam \\
thang.tranngoc@hust.edu.vn}
}

\maketitle
\thispagestyle{plain}
\pagestyle{plain}

\begin{abstract}
Despite notable results on standard aerial datasets, current state-of-the-arts fail to produce accurate building footprints in dense areas due to challenging properties posed by these areas and limited data availability. In this paper, we propose a framework to address such issues in polygonal building extraction. First, super resolution is employed to enhance the spatial resolution of
aerial image, allowing for finer details to be captured. This enhanced
imagery serves as input to a multitask learning module, which consists
of a segmentation head and a frame field learning head to effectively
handle the irregular building structures. Our model is supervised by adaptive loss weighting, enabling extraction of sharp edges and fine-grained polygons which is difficult due to overlapping buildings and low data quality. Extensive experiments on a slum area in India that mimics a dense area demonstrate that our proposed approach significantly outperforms the current state-of-the-art methods by a large margin.
\end{abstract}


\section{Introduction}

Extracting building footprints from aerial or satellite images is an important task in remote sensing and computer vision. It is used for urban planning, disaster management, and environmental monitoring. Thanks to the advances of deep learning techniques, promising results have been shown in delineating building footprints from aerial imagery. The initial approaches primarily revolved around the prediction of vectorized building instances, which is often coupled with post-processing methods. Zhao \textit{et al.} \cite{Zhao2018BuildingEF} introduced a multi-step boundary regularization technique using the paradigm of Mask-RCNN, while Li \textit{et al.} \cite{Li2020ApproximatingSI} refined polygonal partitions using U-Net based architectures. A Generative Adversarial Network (GAN) based method was proposed to enhance the quality of the output segmentation without depending excessively on segmentation results \cite{zorzi2021machine}. Later methods involved the direct prediction of building polygon vertices using Convolutional Neural Network  \cite{li2019topological} or Graph Convolutional Networks \cite{ling2019fast}. Recently, frame field learning has been proposed \cite{Girard2020} to capture local orientation and directional attributes from building structures.

While comparable results have been reported on current benchmarks \cite{maggiori2017dataset, crowdai} that cover regular areas, little work has been studied on extraction of buildings in dense areas (e.g. rural slum areas). These areas contain densely packed and irregularly arranged buildings, posing difficulties for existing methods to effectively extract building footprints. 

In this work, we leverage our curated dataset covering a slum area in India, and use it as our benchmark. We propose a novel framework that integrates super resolution, frame field learning, and polygonization techniques, which enhances accuracy and robustness for building outlines extraction in these complex settings. First, super resolution (SR) is employed to capture finer details from complex building structures present in dense areas. The multitask learning module then takes SR imagery as input to produce a segmentation mask and a frame field, which complementarily optimize each other to effectively handle the irregular building outlines. Finally, a polygonization module is utilized to refine the contours, resulting in regular footprints with sharp edges. 

In summary, we propose to enhance the extraction of building footprints in highly dense areas by: (1) incorporating advanced spatial resolution enhancement for aerial imagery (2) combining segmentation with frame field in a multitask learning module and (3) performing polygonization to address the challenges in extracting building footprints in dense areas. Experiments carried out on our benchmark show superiority of our framework over current state-of-the-art methods.

\section{Related Works}
\begin{figure}[t]
  \begin{center}
  \includegraphics[width=\columnwidth]{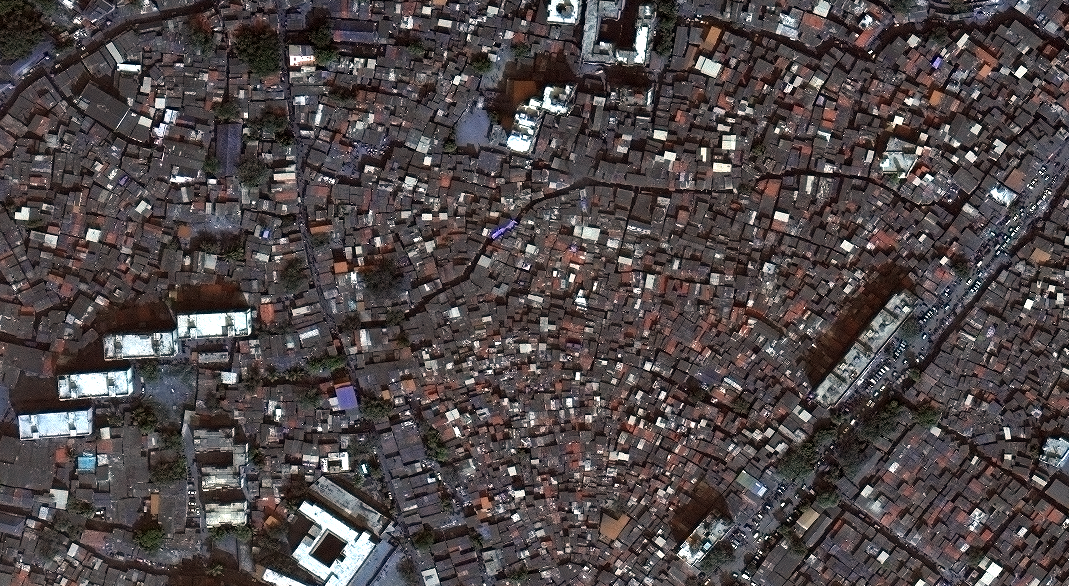}\\
 \caption{Visualization of a part of the aerial image used in this paper. A slum area in Mumbai, India containing around $15,000$ building objects is covered. The study area is characterized with highly dense and irregular building structure.}\label{image_dataset}
  \end{center}
\end{figure}
\subsection{Building Footprint Segmentation}

% SR on aerial images, cite the definition

Building footprint segmentation is the process of identifying and delineating the outlines of buildings from satellite or aerial imagery. Early approaches emphasizes the use of algorithms such as graphs and regions \cite{Bignone1996AutomaticEO}, \cite{10.1007/978-3-319-10599-4_21}. Recent years have seen the dominance of deep learning-based methods in building footprint extraction task \cite{Anh2022, Yu2023, Wahidya2023}. Li \textit{et al.} \cite{Li2020ApproximatingSI} proposed a polygonal partition refinement method for vectorizing the output probability maps
of a U-Net based model. Besides, DeepLabv3+ \cite{chen2017rethinking} leverages dilated convolutions and a decoder module to achieve high-resolution predictions. 
A boundary regularization approach consisting of multiple steps to regulate the predicted building instances generated by Mask-RCNN is introduced in \cite{Zhao2018BuildingEF}, resulting in simplified building polygons.
However, the data dependency problem of the above methods significantly degrades their performance in dense areas which poses severe difficulties for accurate extraction.

\subsection{Super Resolution}

Super resolution method is a technique of recovering a high-resolution image from a low-resolution image. Dong \textit{et al.} \cite{dong2015image} introduced SRCNN which surpassed the performance of previous studies. VDSR \cite{7780551} and IRCNN \cite{zhang2017learning} proposed to increase the network depth by incorporating additional convolutional layers using residual learning. 
Real-ESRGAN \cite{wang2021realesrgan}, which incorporates GAN technique to effectively upscale spatial resolution in aerial imagery, is utilized as one of the key components of our framework.

\subsection{Frame Field Learning}
Frame field learning refers to the process of incorporating a frame field output into a fully-convolutional network, which improves segmentation performance by enhancing sharp corners and provides valuable information for vectorization. It is early used by \cite{Kass2004SnakesAC} to devise a straightforward polygonization. In addition, frame field learning is applied in solving road segmentation tasks \cite{ding2020diresnet}. Girard \textit{et al.} \cite{Girard2020} leverages frame field learning to provide structural information for polygonal building segmentation. However, achieving good generalization across dense areas in real-world scenarios remains a challenge for these methods, where irregular building structures severely undermine the contribution of frame field to improve segmentation results.

\section{Study Area and Dataset}

In this paper, we cope with highly dense areas, which pose
several challenges to the extraction of building contours due to
their properties. First, dense areas are characterized by irregular and ambiguous building patterns along with variations in building types and shapes, which makes defining consistent criteria for extraction not easy. Second, high density leads to overlapping structures and shared walls, making it difficult to distinguish
individual footprints. Last but not least, data sources covering these areas are limited or of low quality.

In this work, we leveraged an aerial image that covers
a slum area of $1,1km^{2}$ in Mumbai, India. The image has a spatial resolution of $0.3m$ as shown in Figure
\ref{image_dataset}. We first converted
the image into three-band, RGB image for further processing. A total
of approximately $15,000$ building objects are present in the
whole study area. We manually annotated the associated building
polygons which are the ground truth contours represented by the sequence
of pixel coordinates that form the outline of the buildings. We then
performed Super Resolution (SR) on the aerial image with details discussed
in the following section. A dataset was then constructed by cropping
the SR aerial image into $1,900$ images of size $256\times256$. We
split the dataset into training, validation and test set with a proportion
of $60\%,20\%$ and $20\%$ respectively. In this study, no data augmentation
method was applied.

\section{Method}

\begin{figure*}
  \begin{center}
  \includegraphics[width=0.99\textwidth]{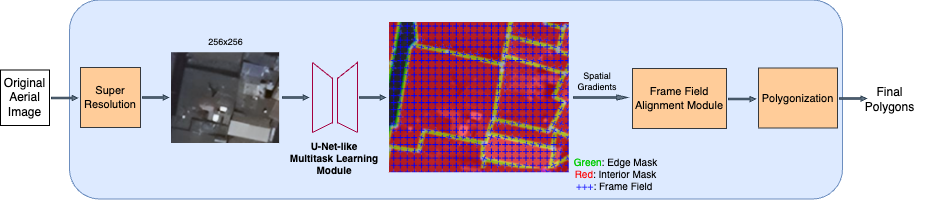}\\
 \caption{Overview of our proposed framework.}\label{baseline}
  \end{center}
\end{figure*}

The overview of our proposed framework is illustrated in Figure \ref{baseline}. First, Super Resolution is performed on the original aerial image. Then, given an image $I$ of size
$3\times256\times256$ as input, the U-Net-like Multi-task Learning module  outputs a segmentation mask and a frame
field. Segmentation masks are then aligned with frame field in Frame Field Alignment module. A Polygonization module is then employed to handle non-trivial topological property of building boundaries.

\subsection{Super Resolution}

Previous frame-field-learning-based works \cite{Girard2021,Xiaoyu2021} have achieved notable results on such datasets covering
standard study areas with sparse and regular building structure. However, these methods fail to extract irregular and dense buildings in dense areas due to low quality and limited data sources. In this work, we propose to apply Super
Resolution to our original aerial image. Enhanced spatial resolution helps
to mitigate the challenges posed by properties of dense areas by boosting
visual sharpness and generating more detailed representations of the
buildings \cite{Shermeyer_2019}. 

We employ Real-ESRGAN \cite{wang2021realesrgan} to enhance spatial
resolution of the original aerial image. Real-ESRGAN is a generative
model that mainly consists of a generator $\mathcal{G}$
and a discriminator $\mathcal{D}$, both are trained in an adversarial
manner. Generator $\mathcal{G}$ reconstructs the corresponding super
resolution (SR) image (denoted as $I_{SR}$) from the given low resolution
(LR) image (denoted as $I_{LR}$), i.e. $I_{SR}=\mathcal{G}(I_{LR})$.
In this work, $I_{LR}$ is the original aerial image (Figure \ref{image_dataset})
with a spatial resolution of $0.3m$, and $I_{SR}$ is upscaled to a spatial
resolution of $0.15m$. Building edges, corners in aerial images of
dense areas are discriminative cues that can aid in accurately
distinguishing buildings from other objects in the scene. The generator
$\mathcal{G}$ restores these texture details by using several
Residual-in-Residual Dense Blocks (RRDB), each combines multi-level
residual network and dense connection with batch normalization removed.
Beside, $\mathcal{G}$ also employs
pixel-unshuffle to reduce the large spatial size while increasing the
channel size, leading to more precise features for footprints extraction.

On the other hand, discriminator $\mathcal{D}$ attempts to distinguish
between real high resolution image and
generated SR image generated by $\mathcal{G}$. The improved U-Net
discriminator $\mathcal{D}$ with skip connections brings a
greater discriminative power for complex aerial image of dense areas. Moreover, discriminator $\mathcal{D}$ provides more detailed
feedback to the generator by outputting pixel-wise realness values.
Hence, the generator is capable of generating sharper local features,
which facilitates the recognition of small and irregular buildings
in dense areas.

\begin{figure}[ht]
     \centering
     \begin{subfigure}[b]{0.49\columnwidth}
         \centering
         \includegraphics[width=\textwidth]{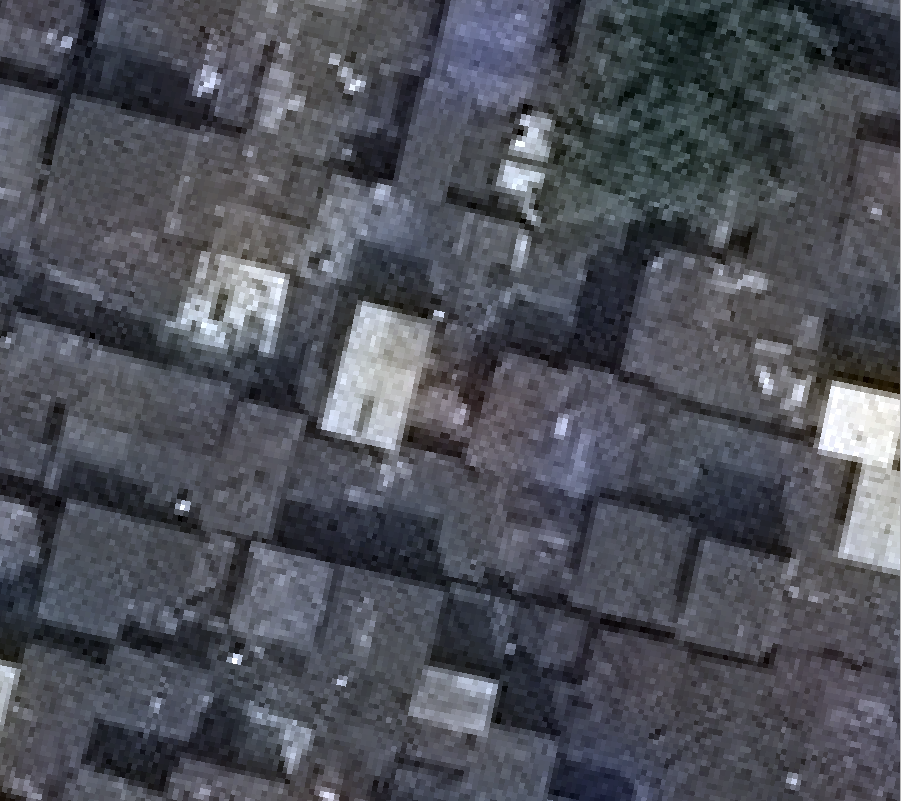}
         \caption{Sample input data}
         \label{sampleinput}
     \end{subfigure}
     \hfill
     \begin{subfigure}[b]{0.49\columnwidth}
         \centering
         \includegraphics[width=\textwidth]{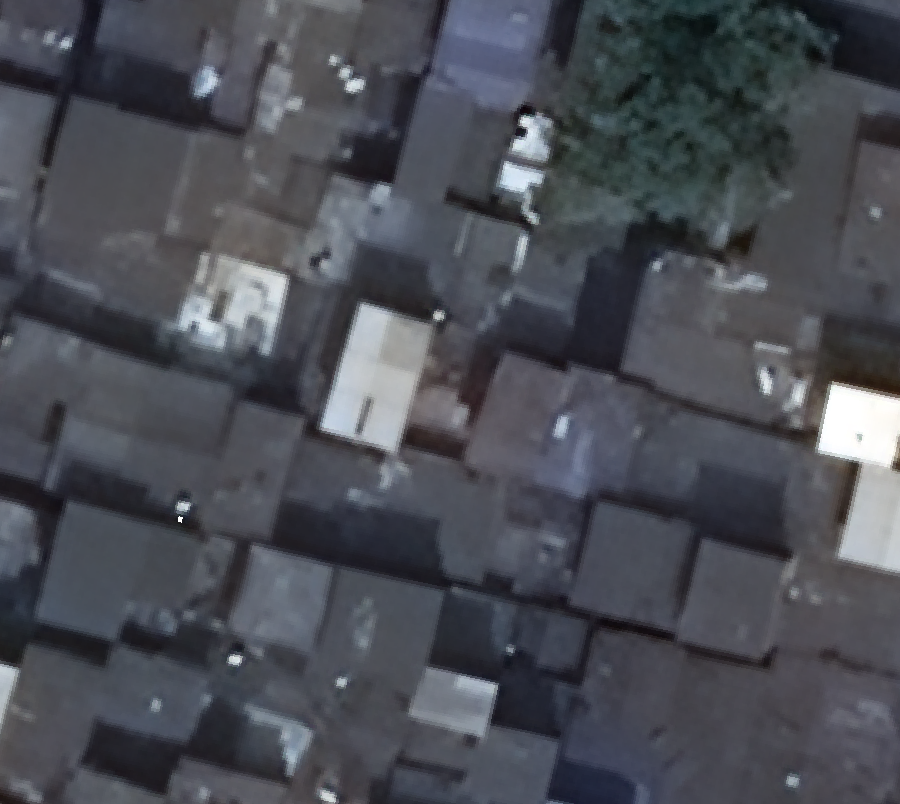}
         \caption{Enhanced SR data}
         \label{SRversion}
     \end{subfigure}
        \caption{SR data enhances details of highly dense building complex for more accurate footprint extraction.}
        \label{SRoutput}
\end{figure}

As shown in Figure \ref{SRoutput}, by enhancing the resolution and quality of the aerial images, we have
enabled the subsequent extraction stage to achieve improved accuracy
in identifying and delineating building footprints. This, in turn,
contributes to better understanding and management of slum areas,
facilitating effective urban planning and development initiatives.

\subsection{Multitask Learning}

\subsubsection{Multitask Learning Module}

Following \cite{Girard2021}, besides outputting segmentation
mask, we leverage Frame Field Learning to accurately extract polygonal
edges and sharp corners without the need for high-cost post-processing
techniques. In a frame field plane, each point is assigned a set of four vectors
$\{i,-i,j,-j\}$. Two complex numbers $i,j\in\mathbb{C}$ define the
directionality at each point, thus, help capture missing shape information
at corners. The pair of directions of each frame can be easily
recovered given a pair $(k_{0},k_{1})$ which are the encoded coefficients
of:
\begin{equation}\label{eq1}
f(z)=(z^{2}-i^{2})(z^{2}-j^{2})=z^{4}+k_{1}z^{2}+k_{0}.    
\end{equation}
A smooth frame field is obtained by learning a pair $(k_{0},k_{1})$ in \ref{eq1}
at each pixel as explored in \cite{bessmeltsev2018vectorization}. Note that the field at polygon
corners aligns to both tangent directions to ensure the frame field
not collapse to a line field.  

As shown in Figure \ref{baseline}, the Multitask Learning
module leverages a U-Net-like network to simultaneously learn the
segmentation masks and a frame field. The network is constructed with
two internal branches, both take an $F$-dimensional feature map $f\in\mathbb{R}^{F\times256\times256}$
as input. One branch produces the segmentation map $f^{seg}\in\mathbb{R}^{2\times256\times256}$
which is a binary classification map with two labels: building interiors
(denoted as $f^{int}$) and building boundaries (denoted as
$f^{edge}$). The other branch combines the segmentation masks
and output feature maps $[f,f^{edge}]\in\mathbb{R}^{(F+2)\times256\times256}$
as input and outputs the frame field with $\hat{k}_{0},\hat{k}_{1}\in\mathbb{C}^{256\times256}$.

Binary cross-entropy loss $\mathcal{L}_{BCE}$ is leveraged jointly with Dice loss $\mathcal{L}_{Dice}$ \cite{Sudre_2017} to drive the segmentation head. Given ground truth $y^{int}$ and $y^{edge}$, interior mask $f^{int}$
is corrected by $\mathcal{L}_{int}$ formulated as:
\begin{equation}\label{intloss}
\mathcal{L}_{int}=c\mathcal{L}_{BCE}(f^{int},y^{int})+(1-c)\mathcal{L}_{Dice}(f^{int},y^{int})
\end{equation}
and corresponding edge mask $f^{edge}$
is corrected by $\mathcal{L}_{edge}$ analogous to \ref{intloss}. In this work, $c$ was set to $0.25$. 
% pixel-wise pair 
% and . Specifically, 
On the other hand, given the ground truth angle $\theta_{\eta}\in[0,\pi)$ of the unsigned tangent vector of the polygon contour, frame field is
forced to by aligned to tangent directions $\hat{k}_{0},\hat{k}_{1}\in\mathbb{C}^{H\times W}$ by the alignment loss:
\begin{equation}\label{alignloss}
\mathcal{L}_{align}=\frac{1}{256^{2}}\sum_{x\in I}y^{edge}(x)|f\left(e^{i\theta_{\eta}};\hat{k}_{0}(x),\hat{k}_{1}(x)\right)|^{2}
\end{equation}
\begin{comment}
\[
\mathcal{L}_{align90}=\frac{1}{256^{2}}\sum_{x\in I}y^{edge}(x)|f(e^{i\theta_{\eta-\frac{\pi}{2}}};\hat{k}_{0}(x),\hat{k}_{1}(x)|^{2}
\]
\end{comment}
while the orthogonal alignment loss $\mathcal{L}_{align90}$ analogous to \ref{alignloss} prevents line-collapsing of frame field by aligning it to
$\eta-\frac{\pi}{2}$. 
% \begin{equation}\label{intloss}
% \mathcal{L}_{int}=c\mathcal{L}_{BCE}(f^{int},y^{int})+(1-c)\mathcal{L}_{Dice}(f^{int},y^{int})
% \end{equation}
\subsubsection{Frame Field Alignment Module}
Given a building, initially, a contour is derived from the segmentation output. Then, the contour is iteratively refined by aligning it with the frame field, ensuring sharp corners with accurate directions. Specifically, first, directionality of frame field is corrected by smoothening $\hat{k}_{0}(x),\hat{k}_{1}(x)$ by $\mathcal{L}_{smooth}$ given as:
\begin{equation}\label{smoothloss}
\mathcal{L}_{smooth}=\frac{1}{256^{2}}\sum_{x\in I}\|\triangledown\hat{k}_{0}(x)\|^{2}+\|\triangledown\hat{k}_{1}(x)|^{2}\|
\end{equation}
Output coupling losses are used to enhance consistency between segmentation
and frame field outputs. Spatial gradients of the $f^{int}$
and $f^{edge}$ are aligned with the frame field by interior
alignment loss $\mathcal{L}_{int\,align}$ and edge alignment loss
$\mathcal{L}_{edge\,align}$ which are analogous to \ref{alignloss}. %
\begin{comment}
\[
\mathcal{L}_{int\,align}=\frac{1}{HW}\sum_{x\in I}f\left(\triangledownf^{int}(x);\hat{k}_{0}(x),\hat{k}_{1}(x)\right)|^{2}
\]
\[
\mathcal{L}_{edge\,align}=\frac{1}{HW}\sum_{x\in I}f\left(\triangledownf^{edge}(x);\hat{k}_{0}(x),\hat{k}_{1}(x)\right)|^{2}
\]
\end{comment}
As buildings in dense areas are closely packed together making them
likely to share common walls, $\mathcal{L}_{int\,edge}$ is built
to improve the model's ability to detect adjoining buildings in the edge mask:

\begin{align} \label{intedgeloss}
\mathcal{L}_{int\,edge} & =\frac{1}{HW}\sum_{x\in I}\max\left(1-f^{int}(x),\|\triangledown f^{int}(x)\|_{2}\right)\\
\, & \cdot|\|\triangledown f^{int}(x)\|_{2}-f^{edge}(x)| \notag
\end{align}
\subsubsection{Adaptive Loss Weighting} 
% The training of the framework is driven by
%  the sum of loss functions $\mathcal{L}_{edge},\mathcal{L}_{int},\mathcal{L}_{align},\mathcal{L}_{align90},\mathcal{L}_{smooth},\mathcal{L}_{int\,align},\mathcal{L}_{edge\,align}$
% and $\mathcal{L}_{int\,edge}$. 
In this work, different from \cite{Girard2021} which assigns equal weighting for every loss in the linear combination of final loss, we propose to
further improve multitask learning by amplifying the contribution
of each loss to its corresponding task \cite{kendall2018multitask}. Specifically, noise parameters $\sigma$ is leveraged as variable in minimizing the final loss:

\begin{align*}
\mathcal{L}= &\frac{1}{2\sigma^{2}}(\mathcal{L}_{smooth}+\mathcal{L}_{int\,align}+\mathcal{L}_{edge\,align}+\mathcal{L}_{int\,edge}) + \\
 & \frac{1}{\sigma}\mathcal{L}_{edge}+\frac{1}{\sigma^{2}}(\mathcal{L}_{int}+\mathcal{L}_{align}+\mathcal{L}_{align90})+\log\sigma
\end{align*}
i.e. each regularizer loss is weighted $\frac{1}{2\sigma^{2}}$ while
each classification loss is weighted $\frac{1}{\sigma^{2}}$, except
for $\mathcal{L}_{edge}$ which accounts for the largest weight to
resolve difficulty in extracting sharp edges caused by tiny and overlapping
buildings. Additional regularizer $\log\sigma$ is added to prevent
$\sigma\rightarrow\infty$. 

\subsection{Polygonization}

The building contours output by Frame Field Learning are then post-processed to smoothen and simplify the corners. Leveraging the interior
map and frame field, a polygonization method is designed to first
extract the initial contour and then optimize the contour by an Active
Contour Model (ACM) \cite{Kass1988}. This helps better align the
edges to the frame field and also produce a more regular building
shape. Polygons with low probabilities are removed, leaving final
polygons.

Specifically, building boundary $f^{edge}$
is first skeletonized, creating a skeleton graph, i.e. the graph of
connected pixels of the skeleton image. Using gradient descend, Active Skeleton Model is conducted in which ACM performs optimization stage which iteratively moves the points on the skeleton graph toward optimal positions. Corner vertices are then detected using the direction information from the frame field. Then, polylines
are obtained by splitting the contour at corner vertices. Polylines
are then further simplified, outputting a polygon with a more regular
shape. 

\begin{comment}
Fig5: visualize polygonization process on a specific building from
a patch
\end{comment}

\section{Experimental setup}

\subsection{Evaluation Metrics}

In this work, for each building object, we
first computed $IoU$ value which measures the intersection area between
predicted segmentation masks and ground truths. F-1 score, Average Precision (AP),
and Average Recall (AR) are measured under the $IoU$ threshold ranging
from $0.50$ to $0.95$ with a step of $0.05$. We report the results under
the $IoU$ thresholds of $0.50$ and $0.75$, giving $F1_{50}$,
$F1_{75}$, $AP_{50}$,
$AP_{75}$, and $AR_{50}$, $AR_{75}$ respectively.
We also report Mean Average Precision (denoted as $mAP$) and Mean
Average Recall (denoted as $mAR$), which provide a comprehensive
understanding of the model's accuracy in accurately
delineating building footprints.
\begin{table*}[ht]
\caption{\label{results}Comparison in quantitative results on our dataset. SR denotes super resolution data, while ORG denotes original data. Overall,
our framework outperforms current state-of-the-art by a large margin especially
with high $IoU$ thresholds. }
\begin{tabular*}{0.99\textwidth}{@{\extracolsep{\fill}} c|cccccccc}
\hline
Method & $F1_{50}$ & $F1_{75}$ & $mAP$ & $mAR$ & $AP_{50}$ & $AP_{75}$ & $AR_{50}$ & $AR_{75}$ \\
\hline\hline
Mask-RCNN\cite{he2018mask}-U$^{2}$Net\cite{Qin2020} + SR & $45.2$ & $22.7$ & $32.1$ & $18.5$ & \textbf{$\mathbf{61.9}$} & $31.2$ & $35.6$ & $17.9$ \\

Multitask Learning \cite{Girard2021} + ORG & $41.5$ & $27.7$ & $26.7$ & $24.2$ & $43.7$ & $29.1$ & $39.6$ & $26.4$ \\

Multitask Learning + SR (\textbf{Ours}) & $\mathbf{59.1}$ & $\mathbf{39.8}$ & $\mathbf{34.5}$ & $\mathbf{38.9}$ & $55.8$ & $\mathbf{37.5}$ & $\mathbf{62.9}$ & $\mathbf{42.3}$ \\
\hline
\end{tabular*}
\end{table*}

\begin{figure*}[ht]
    \centering

    \begin{subfigure}{0.24\textwidth}
        \includegraphics[width=\linewidth]{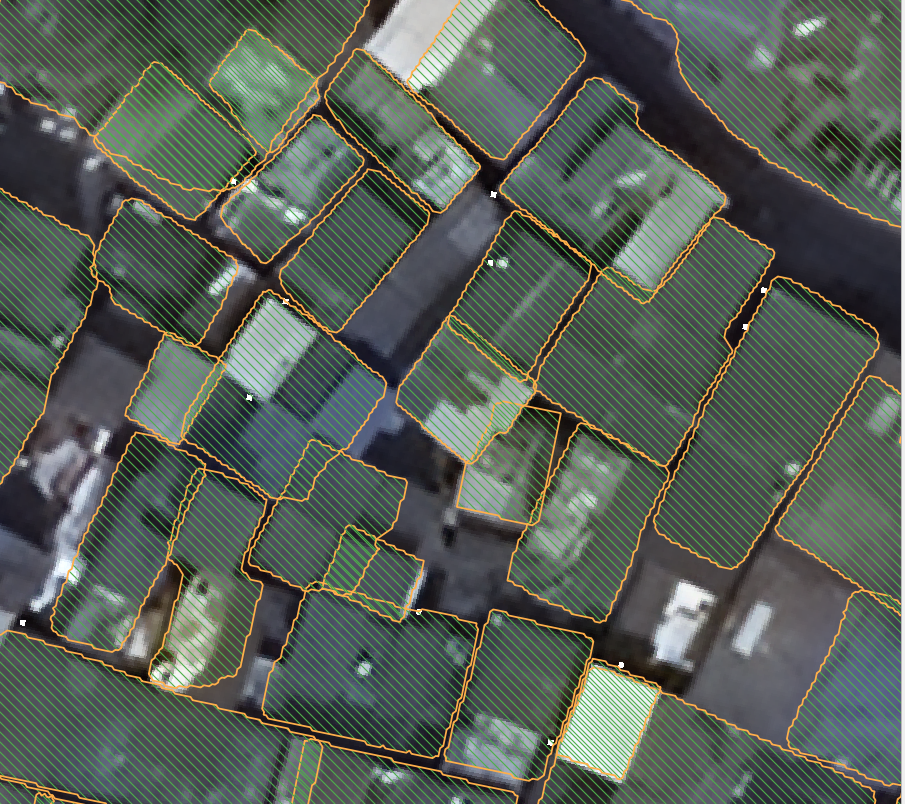}
    \end{subfigure}
    \hfill
    \begin{subfigure}{0.24\textwidth}
        \includegraphics[width=\linewidth]{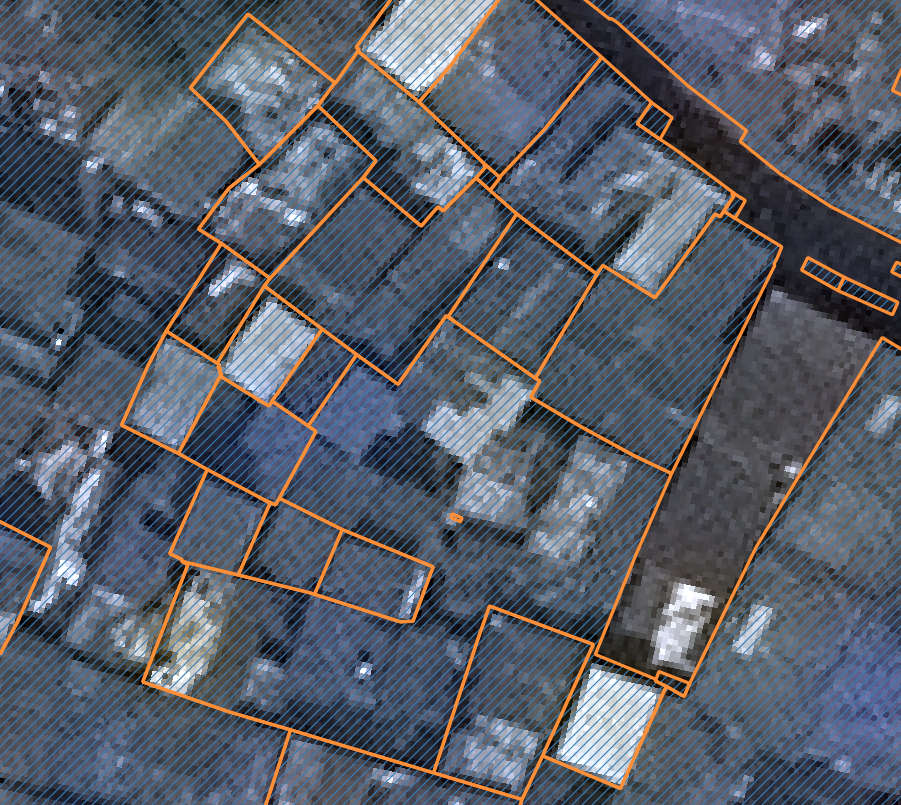}
    \end{subfigure}
    \hfill
    \begin{subfigure}{0.24\textwidth}
        \includegraphics[width=\linewidth]{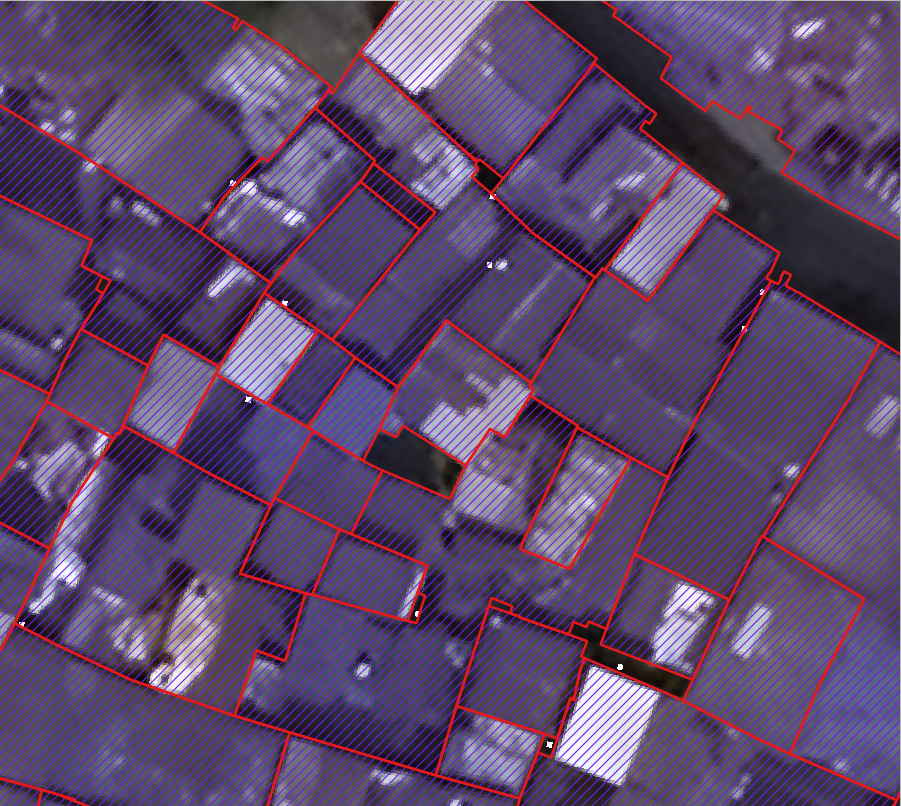}
    \end{subfigure}
    \hfill
    \begin{subfigure}{0.24\textwidth}
        \includegraphics[width=\linewidth]{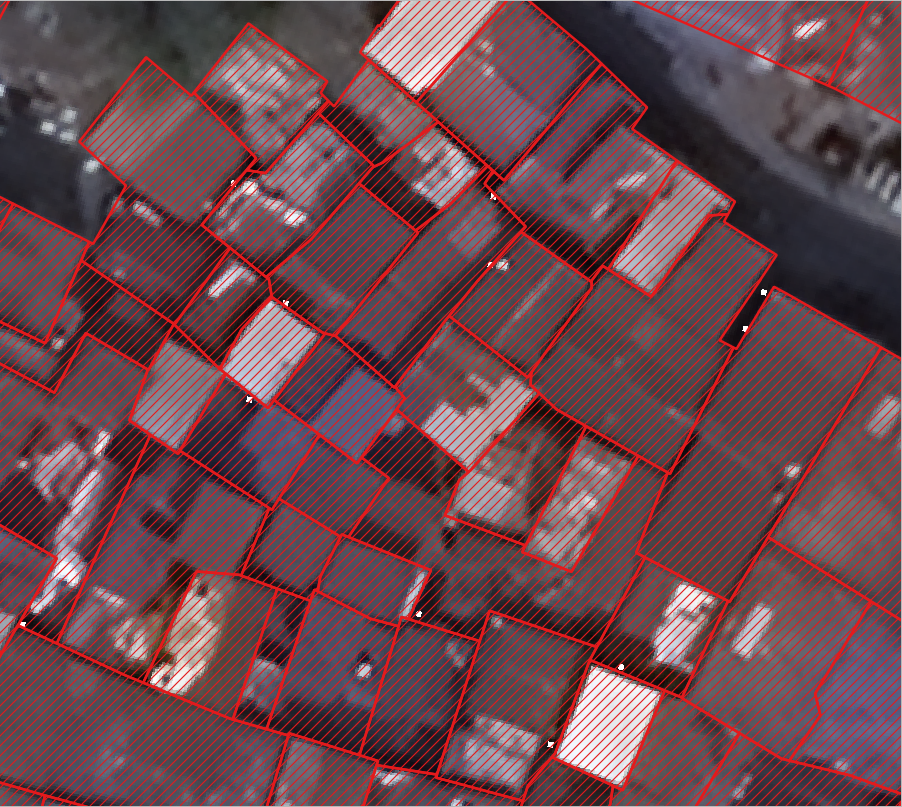}
    \end{subfigure}

    \vspace{0.3cm}

    \begin{subfigure}{0.24\textwidth} 
        \includegraphics[width=\linewidth]{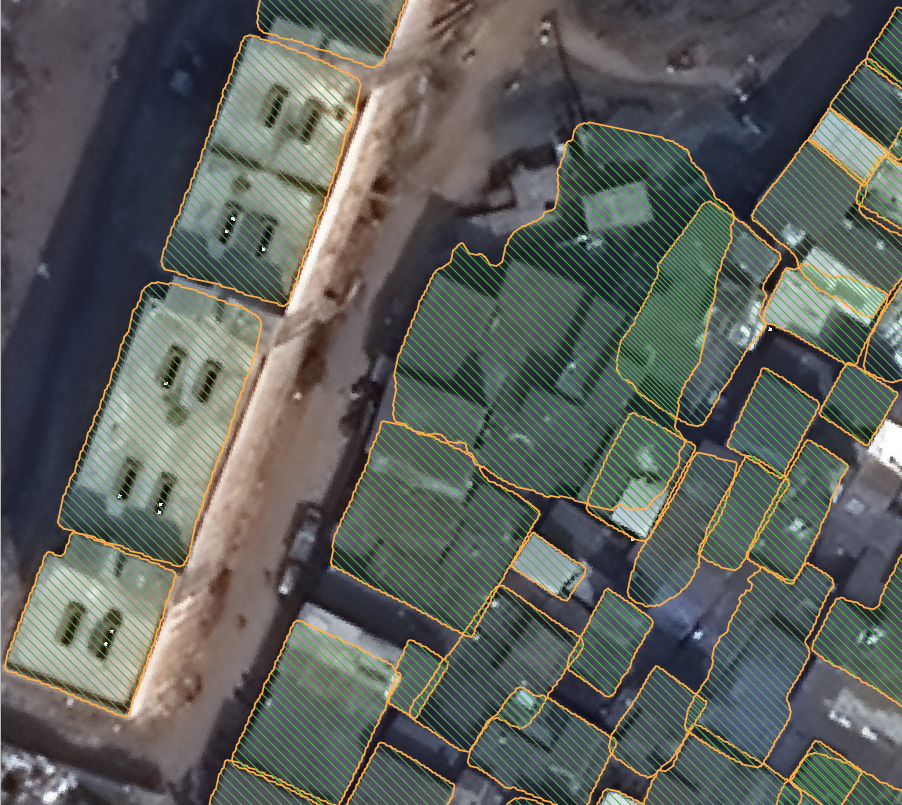}
        \caption{\label{mrcnn_sr_results}MaskRCNN-U$^{2}$Net + SR}
    \end{subfigure}
    \hfill
    \begin{subfigure}{0.24\textwidth} 
        \includegraphics[width=\linewidth]{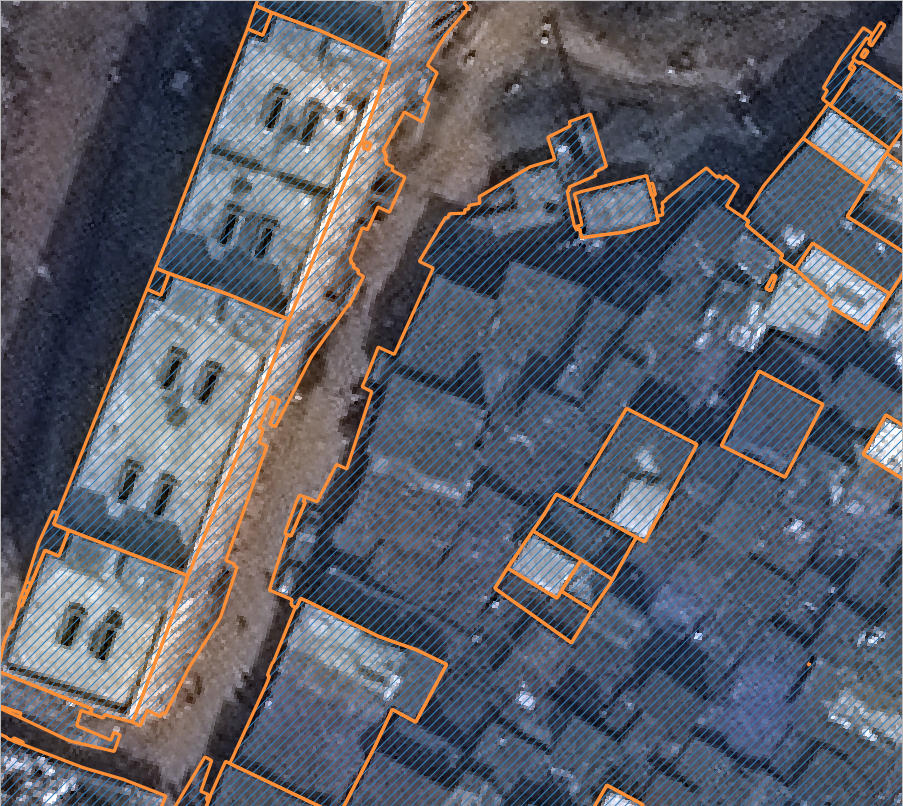}
        \caption{\label{ffl_org_results}Multitask Learning + ORG}
    \end{subfigure}
    \hfill
    \begin{subfigure}{0.24\textwidth} 
        \includegraphics[width=\linewidth]{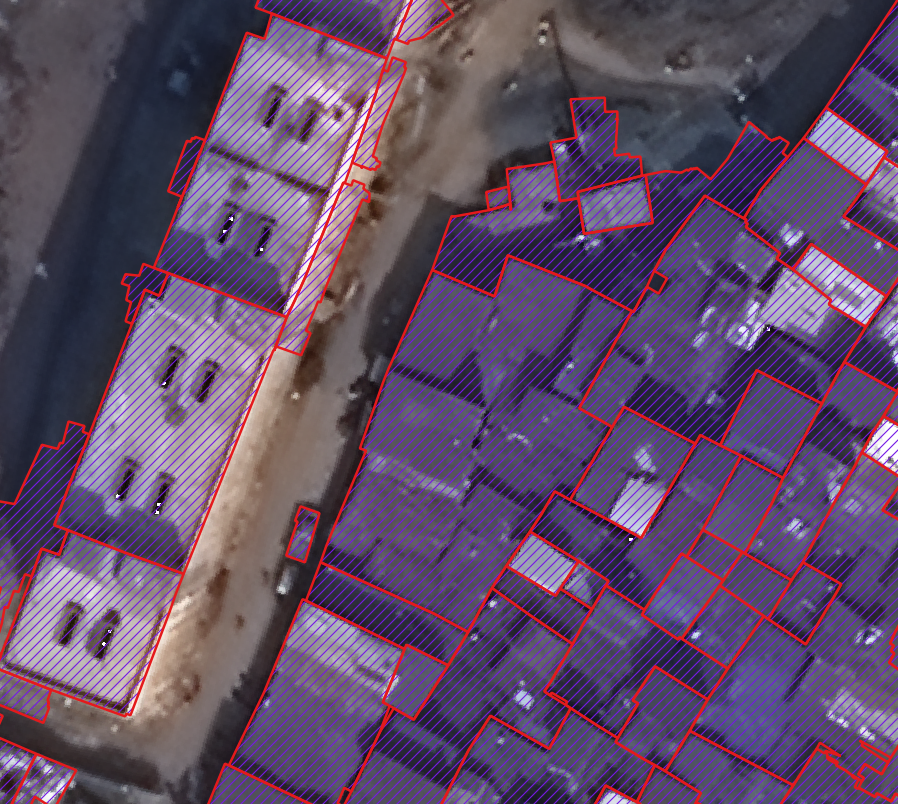}
        \caption{\label{proposed_results}\textbf{Our proposed framework}}
    \end{subfigure}
    \hfill
    \begin{subfigure}{0.24\textwidth} 
        \includegraphics[width=\linewidth]{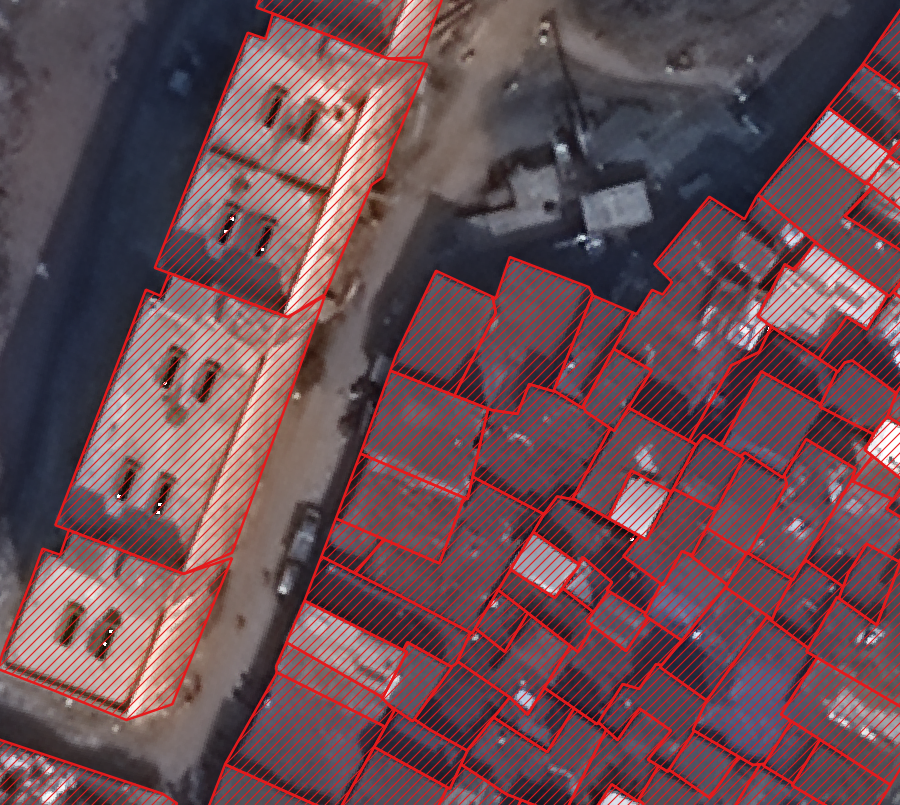}
        \caption{Ground truth}
    \end{subfigure}

    \caption{\label{extraction_results}The qualitative comparison in visualized outputs among Mask-RCNN-U$^2$Net on super resolution data (SR), Frame Field Learning on original data (ORG) and our proposed framework.}
\end{figure*}

\subsection{Implementation Details}

For Super Resolution, we leveraged RealESR-GAN \cite{wang2021realesrgan},
where we pretrained it with $6120$ aerial images with a scale list of
$[1,0.75,0.5,1/3]$. The images containing mostly building objects
were gathered from several datasets which cover a broad range of spatial
resolution (from $0.04m$ to $0.5m$) and multiple areas to improve generalization ability.
Training Real-ESRGAN is conducted with $100,000$
iterations. The pretrained model is then used to perform inference on our
aerial image (Figure \ref{image_dataset}). The output SR image is then
cropped to create input images for the subsequent Multitask Learning
module. 

For Multitask Learning module backbone, we used a DeepLabV3 \cite{chen2017rethinking}
model called DeepLab101, in which a ResNet-101 encoder \cite{he2015deep}
is utilized. Instead of random initialization, we employed pre-trained
model from \cite{Girard2021}. Following the original U-Net \cite{ronneberger2015unet},
distance weighting for the cross-entropy loss is also used. We trained our model with $200$
epochs using Adam optimizer with a batch size of $16$, an initial
learning rate of $0.001$ and a decay rate of $0.9$.
\section{Results and Discussion}

Quantitative results on our dataset (Figure \ref{image_dataset}) are reported in Table \ref{results}, while qualitative extraction results are visualized in Figure \ref{extraction_results}. We compare our proposed
framework with two state-of-the-arts (SOTAs) including the segmentation-based
method Mask-RCNN \cite{he2018mask} with backbone U$^{2}$-Net
\cite{Qin2020} trained on Super Resolution (SR) data and the multitask-learning-based
method proposed in \cite{Girard2021} trained on original data. Overall, our proposed framework outperforms previous SOTAs by a significant margin in all evaluation metrics.

\subsubsection{Comparison with Semantic Segmentation Model on Super Resolution data}

From Table \ref{results}. Our method outperforms Mask-RCNN-U$^{2}$Net in $mAR$ by a large margin of $20.4\%$. The performance of our method is notably more remarkable than Mask-RCNN-U$^{2}$Net in both Average Precision and Average Recall on high $IoU$ thresholds,
i.e. $17.1\%$ higher in $F1_{75}$, $27.3\%$ higher in $AP_{75}$, and $24.3\%$ higher
in $AR_{75}$. Significant improvement on high IoU threshold indicates more precise extraction
results produced by our proposed framework compared to the segmentation-based SOTA method. 

\begin{figure}[ht]
     \begin{subfigure}[b]{0.49\columnwidth}
         \centering
         \includegraphics[width=\columnwidth]{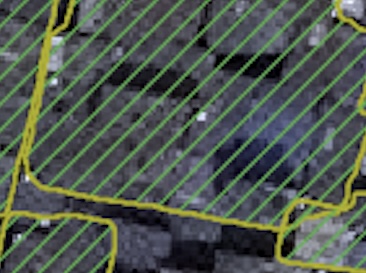}
         % \caption{MRCNN-U$^2$Net failure case}
         \label{Mask-RCNN-failure}
     \end{subfigure}
     \hfill
     \begin{subfigure}[b]{0.49\columnwidth}
         \centering
         \includegraphics[width=\columnwidth]{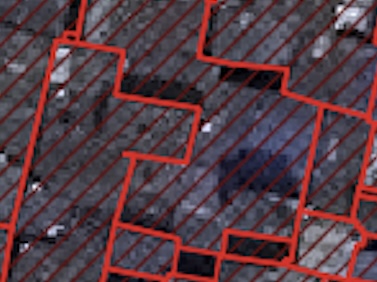}
         % \caption{Our proposed framework}
         \label{proposed-success}
     \end{subfigure}
     
        \caption{A failure case of Mask-RCNN-U$^2$Net (left) compared to our proposed framework (right) on original data.}
        \label{comp-Mask-RCNN}
\end{figure}

The qualitative comparison on extraction results visualized in Figure \ref{mrcnn_sr_results} and 
\ref{proposed_results} demonstrate that our proposed framework is superior to Mask-RCNN-U$^{2}$Net
in many aspects. First, more complete building outlines are obtained. Second, tiny and neighboring buildings
from a highly dense region are accurately extracted, compared to false sparse building outlines produced by Mask-RCNN-U$^{2}$Net. In Figure \ref{comp-Mask-RCNN}, we show a
failure case of Mask-RCNN-U$^{2}$Net and how our proposed framework effectively
handled the tiny and overlapping buildings in dense areas. The building
boundaries are also well-refined using our framework, shown by more
regular polygons with thinner, smoother edges and sharper corners. The reason is Mask-RCNN-U$^{2}$Net solely relies on the ability of model to detect bounding boxes and
segment objects from bounding boxes, which is severely affected by
overlapping masks. On the other hand, by combining frame field and segmentation, our proposed framework simultaneously obtain masks
and contours, followed by a frame-field-based polygonization stage
to better align polygon edges and vertices.

\subsubsection{Effectiveness of Super Resolution and Adaptive Loss Weighting}
To demonstrate that the multitask learning is improved by utilizing Super Resolution (SR), we also compare our proposed framework with the method proposed in \cite{Girard2021} trained on original data, while we trained our method on SR data with adaptive
loss weighting. From Table \ref{results}, it can be seen that our proposed
framework is superior in all evaluation metrics,
i.e. $7.8\%$ higher in $mAP$ and $14.7\%$ higher in $mAR$, showing
that the proposed Super Resolution stage significantly improves the
ability of model to learn accurate frame field and regular polygons,
while building edges are well captured by adaptive loss weighting.

\begin{figure}[H]

    \begin{subfigure}{0.49\columnwidth}
        \includegraphics[width=\linewidth]{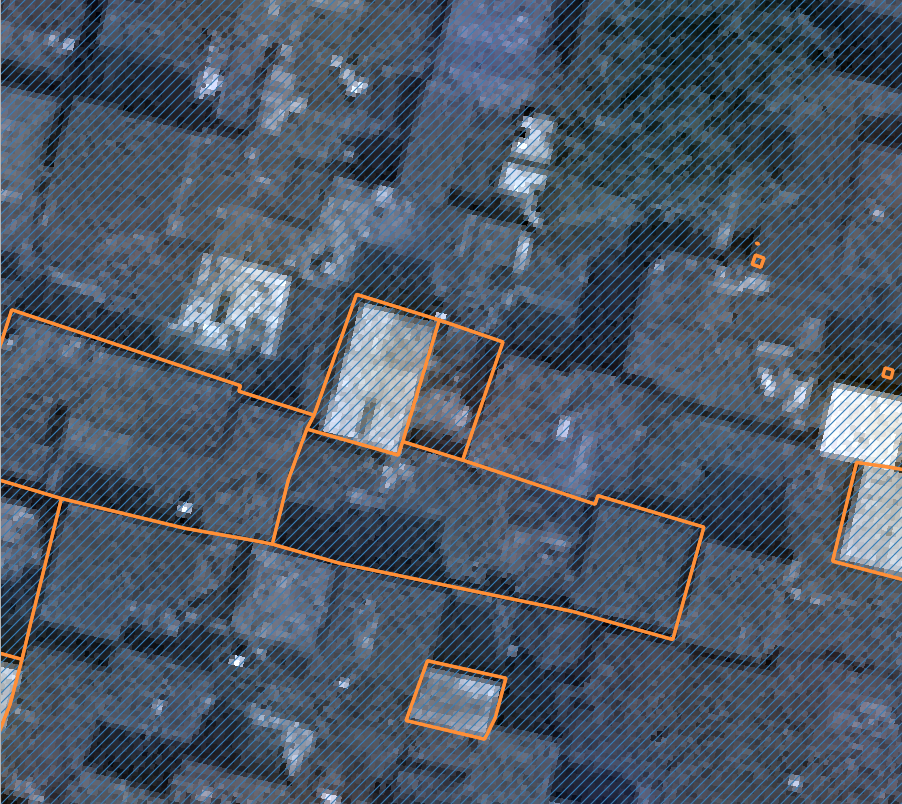}
        % \caption{Subfigure 6}
    \end{subfigure}
    \hfill
    \begin{subfigure}{0.49\columnwidth}
        \includegraphics[width=\linewidth]{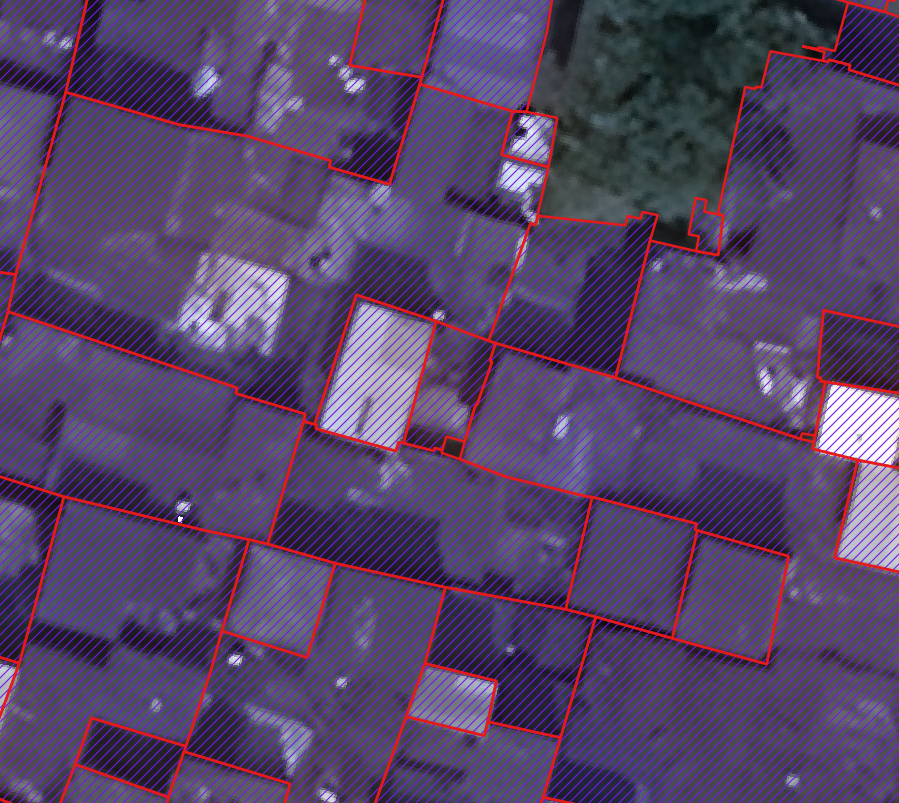}
        % \caption{Subfigure 7}
    \end{subfigure}

    \caption{A failure case of the method proposed in \cite{Girard2021} trained on original data (left). Our method trained on super resolution data shows superiority and robustness in extraction (right).}
    \label{failure_ffl_org}
\end{figure}

In terms of qualitative extraction results, it can be seen from Figure \ref{ffl_org_results}
and \ref{proposed_results} that low image quality remarkably degrades the performance of the framework in \cite{Girard2021}, which suffers from severe confusion caused by blur edges and high density. Using SR, the segmentation-based Mask-RCNN-U$^{2}$Net is able to produce more complete building outlines than the multitask-learning-based method in \cite{Girard2021}. The failure case visualized in Figure \ref{failure_ffl_org} demonstrates that enhanced spatial resolution of the aerial image enables our method to recognize ambiguous contours even for tiny buildings. Edges are further sharpened, while invalid vertices are filtered out, which makes our model robust to poor segmentation results.

\section{Conclusion}

In this paper, we have addressed the challenges encountered in extracting building outlines from aerial images of dense areas by employing the techniques of Super Resolution (SR) and Multitask Learning. Extensive experiments demonstrate that utilizing SR to enhance the spatial resolution of aerial image improves the accuracy of the subsequent segmentation process. Moreover, the integration of frame field learning facilitated the preservation of local textures and fine details crucial for the extraction of sharper edges, which surpasses existing techniques in terms of accuracy. The superior performance achieved by our proposed approach highlights its potential and future directions for practical applications in urban planning, which aim at improving living conditions in slum areas. 
\bibliographystyle{IEEEtran}
\bibliography{refs}

\end{document}